\documentclass[journal]{IEEEtran}
\usepackage{lipsum} 
\usepackage[tight,footnotesize]{subfigure}
\usepackage{enumitem} 
\usepackage{float}
\usepackage{multicol,multienum}
\usepackage{breqn}
\usepackage{stfloats}
\usepackage{multirow}
\usepackage{diagbox}
\usepackage{hyperref}
\usepackage{bm}
\usepackage{cite}
\usepackage{graphicx}
\usepackage{tabularx}
\usepackage[ruled,vlined]{algorithm2e} 
\usepackage{algorithmic}
\usepackage{booktabs}
\usepackage{amssymb}
\usepackage{amsmath}
\usepackage{amsthm}
\usepackage{thmtools}
\newtheorem{lemma}{Lemma}

\newtheorem{theorem}{Theorem}
\newtheorem{assumption}{Assumption}
\usepackage[tight,footnotesize]{subfigure}
\usepackage{url}
\usepackage{doi}
\usepackage{color}

\newenvironment{sloppypar*}
{\sloppy\ignorespaces}
{\par}

\ifCLASSINFOpdf

\else

\fi

\begin{document}\sloppy

\title{Resource-constrained Federated Edge Learning with Heterogeneous Data: Formulation and Analysis}

\author{Yi Liu\textsuperscript{$\dagger$},
        Yuanshao Zhu\textsuperscript{$\dagger$},
        and James J.Q. Yu,~\IEEEmembership{Senior Member,~IEEE}
\thanks{A preliminary version of this paper which has the title ``FedOVA: One-vs-All Training Method for Federated Learning with Non-IID Data” was presented at IEEE IJCNN 2021 \cite{9533409}. Yi Liu was with the Guangdong Provincial Key Laboratory of Brain-inspired Intelligent Computation, Department of Computer Science and Engineering, Southern University of Science and Technology, Shenzhen 518055, China. He is now with the Department of Computer Science, City University of Hong Kong, Hong Kong SAR 518057, China (e-mail: 97liuyi@ieee.org). Yuanshao Zhu and James J. Q. Yu are with the Guangdong Provincial Key Laboratory of Brain-inspired Intelligent Computation, Department of Computer Science and Engineering, Southern University of Science and Technology, Shenzhen 518055, China (e-mail: yasozhu@gmail.com; yujq3@sustech.edu.cn).

\textsuperscript{$\dagger$}Equal contributions and James J.Q. Yu is the corresponding author.}}

\markboth{IEEE Transactions on Network Science and Engineering}%
{Shell \MakeLowercase{\textit{et al.}}: Bare Demo of IEEEtran.cls for IEEE Journals}

\maketitle

\begin{abstract}
Efficient collaboration between collaborative machine learning and wireless communication technology, forming a Federated Edge Learning (FEEL), has spawned a series of next-generation intelligent applications. %
{However, due to the openness of network connections, the FEEL framework generally involves hundreds of remote devices (or clients), resulting in expensive communication costs, which is not friendly to resource-constrained FEEL. To address this issue, we propose a distributed approximate Newton-type algorithm with fast convergence speed to alleviate the problem of FEEL resource (in terms of communication resources) constraints. Specifically, the proposed algorithm is improved based on distributed L-BFGS algorithm and allows each client to approximate the high-cost Hessian matrix by computing the low-cost Fisher matrix in a distributed manner to find a ``better'' descent direction, thereby speeding up convergence. Second, we prove that the proposed algorithm has linear convergence in strongly convex and non-convex cases and analyze its computational and communication complexity.} {Similarly, due to the heterogeneity of the connected remote devices, FEEL faces the challenge of heterogeneous data and non-IID (Independent and Identically Distributed) data.} To this end, we design a simple but elegant training scheme, namely FedOVA, to solve the heterogeneous statistical challenge brought by heterogeneous data. In this way, FedOVA first decomposes a multi-class classification problem into more straightforward binary classification problems and then combines their respective outputs using ensemble learning. {In particular, the scheme can be well integrated with our communication efficient algorithm to serve FEEL. Numerical results verify the effectiveness and superiority of the proposed algorithm.}
\end{abstract}


\IEEEpeerreviewmaketitle

\section{Introduction}
\IEEEPARstart{W}{ith} the convergence of machine learning technology and wireless communication technology, various smart mobile applications have been developed to provide people with high-quality services, which have revolutionized our daily lives and brought benefits to the human societies and national economy \cite{9045112}. 
The key idea of developing machine learning-based smart mobile applications in wireless networks is to learn (train) models leveraging the rich data collected by massively distributed user devices \cite{kang2020scalable}. 
{For example, traditional cloud computing-based centralized machine learning frameworks focus on central data processing, which generally requires widely distributed mobile devices to offload their local training data to remote cloud centers for centralized learning} \cite{9252924,9127160}. 
Nevertheless, such a solution still faces two severe challenges: limited communication resources and data privacy. It is impractical for edge devices with limited resources (in terms of computation and communication resources) to offload massive amounts of local training data to the remote cloud center. 
On the other hand, since the remote cloud center is inevitably attacked by some malicious opponents, the private information involved in the training data uploaded by the edge device may be leaked \cite{liu2020towards}. 
Therefore, such a framework is no longer suitable for a modern society that values the right to privacy, motivating us to develop a new learning framework to solve the above challenges.

{In light of the two challenging issues, the researchers designed a novel architecture called Mobile Edge Computing (MEC) \cite{mao2017survey}, which significantly reduces communication latency and mobile energy consumption by migrating cloud computation capability and learning tasks from remote cloud center to edge servers \cite{luo2020cost,9346022}. Unfortunately, previous work cannot be used to mitigate our challenging issues. The main reason is that the server in the MEC architecture still needs to access the original data of the remote device, which still leads to privacy leaks \cite{liu2020federated}.}
To address the privacy concern, a collaborative model training scheme was proposed, i.e., Federated Learning \cite{mcmahan2017communication}, which leverages large amounts of data distributed over multiple edge devices, such that the latter do not share locally private data with any centralized entity. 
Concretely, edge devices (or \emph{clients}) are asked to upload model updates to a centralized server iteratively, thereby jointly training a shared global model.
In this context, the central server can use the rich distributed data to train the optimal shared model and can rationally allocate the computation resources of the edge devices \cite{9252924}. 
Inspired by the above facts, effective collaboration between MEC and FL (as known as Federated Edge Learning (FEEL)) has great potential to promote the use of collaborative machine learning in next-generation smart applications.

Although the FEEL framework has achieved great success in edge computing networks and spawned a series of emerging applications such as Google Keyboard \cite{leroy2019federated}, it still suffers from heterogeneous data and resource constraints ({in terms of communication cost}) \cite{dinh2020done}. 
Unlike the MEC architecture, the data at the edge is generally heterogeneous and non-independent, and identically distributed (non-IID) in the FEEL framework, which impairs the performance of the machine learning model \cite{9155494,9060868,9359187}. 
{Obviously, due to data heterogeneous challenges and non-IID data issues, resulting in slower convergence speed (i.e., more training rounds), expensive communication overhead is still the bottleneck of FEEL. Therefore, it is necessary to build a bridge to link training schemes that relieve heterogeneous data with efficient communication algorithms.}

{Building such a bridge inevitably needs to answer the following question: \emph{how to develop a distributed training algorithm which uses a minimal number of communication rounds to achieve model performance comparable to that of the benchmark FL in \cite{mcmahan2017communication} for convergence and handles non-IID and heterogeneous data across edge devices?}}
{To answer the above question, our goal is to develop a distributed optimization algorithm to achieve faster convergence while being robust to non-IID data. Towards the goal, we present an (approximate) Newton-type method-based distributed second-order optimization algorithm. Our starting point is leveraging gradient and curvature information to find a ``better” descent direction, thereby significantly speeding up the convergence speed. Standing on top of it, to address non-IID issues, we propose a simple but elegant training scheme, namely FedOVA, which can work effectively with the non-IID data in FEEL by decomposing the multi-classification task into multiple binary classification tasks. Specifically, our insight is to decompose a federated multiclass classification task on non-IID data into multiple binary classification tasks on the client-side by introducing the One-vs-All \cite{10.5555/1005332.1005336} training scheme. }In particular, the main contributions of this paper are summarized as follows:
\begin{itemize}
    \item \textbf{\textit{A communication-efficient distributed approximate Newton-type method for resource-constrained federated edge learning:}} We propose a communication-efficient (in terms of convergence speed) L-BFGS optimization algorithm based on (approximate) second-order information with stochastic batches for the federated edge learning framework, as a novel approach to the empirical risk minimization problems. 
    The classic L-BFGS with stochastic batches is generally unstable, which is not friendly for us to implement this algorithm in a distributed manner. 
    For this reason, we use smooth estimation to evaluate the gradient difference while utilizing the Fisher Information Matrix (FIM) to approximate the Hessian matrix to accelerate the convergence.
    \item \textbf{\textit{A simple but elegant training scheme with heterogeneous data for federated edge learning:}} We design a simple but elegant training scheme, namely FedOVA, to address the heterogeneous data and non-IID problems in FEEL.
    Different from existing solutions, FedOVA addresses this problem from the perspective of learning. Specifically, FedOVA (1) improves performance and convergence speed (2) without introducing additional overhead or operations to the FEEL framework. 
    \item \textbf{\textit{Theoretical properties and convergence analysis:}}  We theoretically comprehensively analyze the convergence of the proposed second-order optimization algorithm and demonstrate the effective cooperation between the Fisher Information Matrix and L-BFGS. 
    Notably, we show that the proposed algorithm can linearly converge to the neighborhood of optimal solutions for convex and non-convex problems under standard assumptions for general empirical risk minimization problems. 
    We replace the Hessian matrix used to approximate the gradient difference with the Fisher Information Matrix in a distributed environment. Such a method is comparable to mainstream first-order optimization algorithms.
    \item \textbf{\textit{Extensive experimentation and model validation:}} Extensive case studies conducted on several public datasets (i.e., F-MNIST, CIFAR-10, and KWS dataset) in computer vision and natural language processing tasks have empirically demonstrated the effectiveness of the proposed method. 
    Experimental results verify that our solution is significantly better than mainstream methods on heterogeneous and non-IID data in terms of classification accuracy and convergence time. 
    In particular, we emphasize that our method is empirically applicable to both convex and non-convex machine learning models.
\end{itemize}

\noindent {\textbf{Remark:} Please note that this work is an extension of the authors' prior conference publication \cite{9533409}. Based on \cite{9533409}, we introduce notable algorithmic and theoretical improvements in this manuscript. Apart from textual amendments and changes, we thoroughly revamp and improve the system design for communication efficient and heterogeneous data in FEEL. Furthermore, we introduce FEEL's FIM-based distributed second-order optimization algorithm to reduce the communication and run-time cost significantly. Lastly, we prove that the proposed algorithm has linear convergence in strongly convex and non-convex cases and analyze its computational and communication complexity. Additional case studies are incorporated to validate the improvements.}

{The rest of this paper is organized as follows. We first review the related work in Sec. \ref{sec-2}. Then, we present some background knowledge of this paper in Sec. \ref{sec-3}. Sec. \ref{sec-4} elaborates on the proposed algorithm. Sec. \ref{sec-4} explains the complexity analysis of the proposed algorithm. We conduct a series of experimental and analyses in Sec. \ref{sec-5}. Finally, we conclude this paper in Sec. \ref{sec-6}.}

\section{Related Work}\label{sec-2}
\subsection{Federated Edge Learning}
Federated Edge Learning (FEEL) is an emerging mobile edge computing framework that integrates the advantages of federated learning (artificial intelligence) and mobile edge computing, which arouses the research interest of researchers. 
The survey \cite{9060868} comprehensively introduced FEEL's architecture, the state-of-the-art techniques, standards, and case studies and gave some future research directions. 
Although there are many unfilled research gaps in the FEEL field, the academic and industrial circles generally focus on topics such as resource optimization \cite{9205252,9127160}, communication efficiency optimization \cite{9415152,9252954}, statistical heterogeneous optimization \cite{9252973,zhao2018federated}, and security and privacy \cite{9109557,liu2020towards}. 
In terms of resource optimization, researchers generally utilized economic tools (e.g., contract theory \cite{kang2019toward}, auction theory \cite{9045112}), and designed some incentive mechanisms \cite{khan2020federated} to implement resource management \cite{9205252}, resource allocation \cite{9127160}, and resource scheduling \cite{shi2020joint} in FEEL. 
For communication efficiency optimization, they utilized compression \cite{haddadpour2021federated} and quantization \cite{reisizadeh2020fedpaq} techniques to reduce the communications volume, thereby improving the communication efficiency of the FEEL framework.
{Furthermore, the FEEL framework's security and privacy research topics have never left the researchers' sight. As for the present, some common malicious attacks against the FEEL framework such as poisoning attacks \cite{tolpegin2020data,fung2020limitations}, backdoor attacks \cite{xie2019dba,bagdasaryan2020backdoor}, adversarial attacks \cite{bhagoji2019analyzing}, and gradient leakage attacks \cite{zhu2020deep} have been extensively studied.}
In the future, more new problems and new challenges will be raised and solved by researchers.

\subsection{Distributed Optimization Methods in Federated Edge Learning}
\subsubsection{First-order distributed optimization methods}
The first-order distributed optimization method is the most common and practical in a distributed learning environment, in which it mainly leverages gradient information to find a descent direction to update the global model iteratively.
Distributed Stochastic Gradient Descent (D-SGD) \cite{8889996} as a classic example of this type of algorithm has derived a series of variants such as ColumnSGD \cite{zhang2020c}, variance-reduced SGD \cite{wu2020federated} and so on. 
However, these algorithms sacrifice communication rounds in exchange for accuracy, which is unfriendly to the resource-constrained FEEL framework. 
And it implies that communication overhead is still a bottleneck problem in the resource-constrained FEEL framework.

\subsubsection{Second-order distributed optimization methods}
{Unlike the classic first-order distributed optimization method, the second-order distributed optimization method utilizes gradient information and second-order information (i.e., curvature information) to find a ``better'' descent direction to update the global model iteratively. }
However, previous work \cite{bottou2018optimization} pointed out that the complex inverse matrix-vector product involved in the calculation of curvature information is not suitable for distributed environments. 
To this end, in \cite{li2019feddane,dinh2020done,wang2018giant}, the authors designed DANE, DONE, and GIANT, respectively, to overcome this problem. 
Specifically, DANE circumvented the challenges mentioned above by designing a well-designed local optimization problem, while DONE proposed a distributed approximate Newton method based on the Richardson iteration. 
GIANT used the harmonic mean Hessian matrix to approximate the true Hessian matrix, significantly reducing communication costs.
{Similar to GIANT, we focus on how to approximate the Hessian matrix to reduce communication costs efficiently. This paper develops a variant of the L-BFGS algorithm that uses the Fisher information matrix to approximate the Hessian matrix to achieve fast convergence.}

\subsection{Non-IID Issues in Federated Edge Learning} 
Statistical challenges (i.e., heterogeneous data and non-IID data) are unresolved issues in the FEEL framework. 
The gradient-based aggregation rules (e.g., FedAvg \cite{mcmahan2017communication}) rely on D-SGD, which is widely used to iteratively train deep learning models under the assumption of IID training data. 
The purpose of learning from IID training data is to ensure that the stochastic gradient is an unbiased estimate of the full gradient \cite{zhao2018federated}. In practice, however, it is unrealistic to assume that the local data on each edge client are always IID. 

Existing works have addressed the non-IID issue in FEEL by designing additional framework mechanisms. For example, Zhao \textit{et al.} in \cite{zhao2018federated,8889996} introduced a data-sharing mechanism to improve FedAvg in non-IID data settings. 
The proposed mechanism involves distributing a small amount of globally shared data containing examples of each class, thereby introducing a trade-off between accuracy and centralization. 
However, this approach unintentionally discloses the client's private data as the dataset is publicly shared, hence violating FL's privacy protection requirement. 
Another popular solution is to design a performance-oriented client selection mechanism. Kopparapu \textit{et al.} in \cite{kopparapu2020fedcd} proposed FedCD, an aggregation method that can clone and delete models to dynamically group devices with similar data, thereby selecting the clients with high-quality updates to mitigate the non-IID issue. 
However, the server needs to calculate the model quality score in each round to decide whether to clone or delete it, introducing additional computational overhead. 
Nonetheless, in real-world applications, applying reinforcement learning within resource-constrained FL environments is likely to be impractical. {However, how to organically combine communication efficient algorithms with non-IID robust algorithms is still an open question.}

\section{Preliminaries}\label{sec-3}
\subsection{Federated Edge Learning Pipeline}\label{sec-2-1}
{In the FEEL setting, we consider a server $\mathcal{S}$ and $K$ clients, participating in training a shared global model $\omega^*$ without sharing their raw private data. In this context, we assume that each client holds a local training dataset ${\mathcal{D}_k} = \{ {x_i},{y_i}\} _{i = 1}^{{n_k}}$, where $n_k$ denotes the number of samples. For the model parameters $\omega  \in {\mathbb{R}^d}$ and a local training dataset $\mathcal{D}_k$, let ${F_k}(\omega ,{x_i})$ be the loss function at the client, and let $f(\omega)$ be the loss function at the server. 
Accordingly, the pipeline of federated learning is defined as follows.}

\noindent \textbf{\textit{Phase 1, initialization:}} First, the server selects a certain proportion $q$ of clients from all clients to participate in an FL learning task. Second, the server broadcasts the initialized global model $\omega_0$ to all clients, i.e., $\omega _0^k \leftarrow {\omega _0}$. 
	
\noindent \textbf{\textit{Phase 2, local training:}} { For the $t$-th training round, each client trains the received global model $\omega _t^k$ on its own local dataset $\mathcal{D}_k$. On the client side, the goal is to minimize the following objective function: ${F_k}(\omega ) = \frac{1}{{{n_k}}}\sum\limits_{i \in {\mathcal{D}_k}} {{\ell_i}(\omega )},$ where $\mathcal{D}_k$ is the set of indexes of data samples on the client $k$. For a given federated learning task (such as image classification task), we typically take cross-entropy loss: ${f_i}(\omega ) = \ell ({x_i},{y_i};\omega ) =  - \sum\limits_{m = 1}^M {({I_{[{y_i} = m]}}\log {\sigma _m}} ({x_i};\omega ))$, where $({x_i};\omega )$ is the Softmax operator and $m$ is the number of classe, as the loss function of the local client, i.e., the loss of the prediction on training example $({x_i},{y_i})$ made with model parameters $\omega$. Furthermore, the gradient and Hessian matrix of $F_k(\omega )$ can be calculated as:
$\nabla F_k(\omega ) = \frac{1}{n}\sum\limits_{i = 1}^n {\nabla {f_i}(\omega )} , H_k(\omega ) = \frac{1}{n}\sum\limits_{i = 1}^n {{\nabla ^2}{f_i}(\omega )}.$}
Then each client uploads its model updates $\Delta {\omega _k} = \eta \nabla {F_k}(\omega )$, where $\eta$ is the learning rate, to the server.
	
\noindent\textbf{\textit{Phase 3, aggregation:}} The server uses a model updates aggregation rule like FedAvg \cite{mcmahan2017communication}  to aggregate all the updates to obtain a new global model $\omega_{t+1}$. Specifically, on the server side, the goal is to minimize the following objective function:
\begin{equation}\label{eq-1}
	f(\omega ) = \sum\limits_{k = 1}^K {\frac{{{n_k}}}{n}{F_k}(\omega ),{F_k}(\omega ) = \frac{1}{{{n_k}}}} \sum\limits_{i \in {\mathcal{D}_k}} {{\ell_i}(\omega )}.
\end{equation}
Note that the above steps will be terminated until the global model reaches convergence.

\subsection{Challenges to Distributed Newton's Method}
In this paper, our goal is to utilize the Newton’s method \cite{li2019feddane,dinh2020done} to efficiently solve the problem of minimizing the empirical risk function in the FEEL framework, thus, we run:
\begin{equation}\label{eq-3}
\begin{aligned}
{\omega _{t + 1}} &= {\omega _t} - {\Big({\nabla ^2}f({\omega _t})\Big)^{ - 1}}\nabla f({\omega _t})\\
&= {\omega _t} - {\Big(\frac{1}{K}\sum\limits_{i = 1}^K {{\nabla ^2}{F_k}({\omega _t})} \Big)^{ - 1}}\Big(\frac{1}{K}\sum\limits_{i = 1}^K {\nabla {F_k}(} {\omega _t})\Big),
\end{aligned}
\end{equation}
where ${{\nabla ^2}{F_k}({\omega _t})}$ and ${\nabla {F_k}(} {\omega _t})$ are respectively the Hessian information ${\mathbf{H}_t} \in {\mathbb{R}^{d \times d}}$ and gradient information computed by each client. To be specific, the server collects and aggregates ${{\nabla ^2}{F_k}({\omega _t})}$ and ${\nabla {F_k}(} {\omega _t})$ uploaded by these clients to update the global model (i.e., update Eq. \eqref{eq-1}). \textit{However, there is a big drawback: the calculation of the Hessian is computationally expensive.} Hence, the core requirement of designing the distributed Newton's method is \textit{Hessian-free communication} and \textit{inverse-Hessian-free computation.} We assume that the size and feature vector dimensions of the client's local dataset are $\mathcal{D}_k$ and $d$, respectively, so that each client needs to send Hessians with size $\mathcal{O}(d^2)$ over the network or compute the inverse Hessian with complexity $\mathcal{O}(\sum\nolimits_{i = 1}^n {{\mathcal{D}_i}} {d^2} + {d^3})$. {This is obviously unrealistic for the resource-constrained FEEL framework. In this paper, to address this challenge, we design an FIM-based distributed L-BFGS method for the resource-constrained FEEL framework to approximate Hessian at a low cost and achieve fast convergence.}

\subsection{One-vs-All Training Scheme} \label{sec:ova}
\begin{figure}[!t]
  \centering
  \includegraphics[scale=0.31]{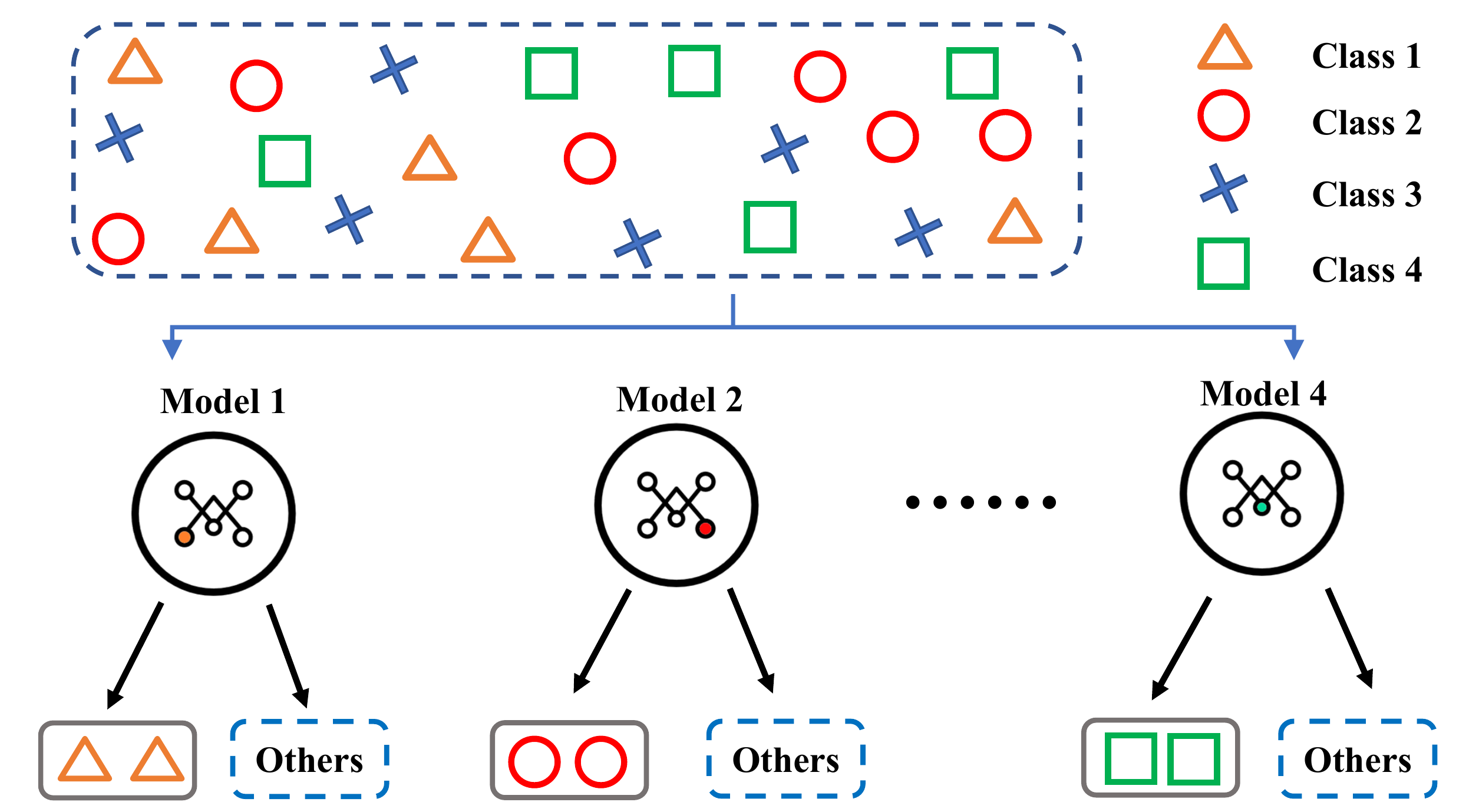}
  \caption{{The OVA training scheme decomposes an $n$-class ($n>2$) classification task into $n$ binary classification tasks, in this figure, $n=4$.}}
  \label{fig:ova}
\end{figure}

One-vs-All (OVA) \cite{rifkin2004defense} training scheme is generally used in logistic regression to solve multi-classification problems.
For example, for an $n$-class ($n>2$) classification task, we assume a labeled dataset $D = \{ {x_i},~{y_i}\} _{i = 1}^n$ where $y_i \in \left\{1,~2,~\dots,~n \right\}$ is the ground-truth label of $x_i$ and $n$ is the total number of samples. As shown in Fig. \ref{fig:ova}, the OVA training scheme trains $n$ binary classifiers ${f_i(\cdot)}, i \in \left\{1,~2,~\dots,~n \right\}$, with each classifier discriminating one class from the others in $D$. Each classifier can be expressed as follows:
\begin{equation}
{f_i}(x) = P(y = i|x;{\omega _i}),
\end{equation}
where $\omega_i$ refers to the parameters of classifier $f_{i}(\cdot)$. To obtain a prediction for a new instance, OVA uses these $n$ binary classifiers to calculate the confidence that the instance belongs to the current class. Specifically, we select the prediction with the highest confidence as the final classification for this new instance. This process is formalized as follows:
\begin{equation}
\hat y = \mathop {\arg \max }\limits_{i \in \left\{1,2,\dots, n \right\}} {f_i}(x).
\end{equation}

\noindent \textbf{Remark:}
It is important to note that, although OVA is a general-purpose approach to multi-class classification that has been used for decades, existing OVA-based works have focused on centralized learning. In this work, we instead leverage distributed OVA to tackle the open problem of distributed multi-class classification under non-IID data.
{Specifically, we take advantage of the independence of each binary classifier in OVA and incorporate it into FEEL to propose a novel training scheme named FedOVA. This scheme is able to solve the non-IID data problem in FEEL efficiently, as will be described in Section \ref{sec-4}. To the best of our knowledge, this is the first work that integrates FEEL and OVA to address FEEL with non-IID data. Moreover, this solution will not affect the normal execution of the communication efficient algorithm, which opens a door for solving the non-IID problem and the expensive communication overhead problem at the same time.}

\begin{table}[!t]
	\centering
	\caption{{Summaries of applied symbols.}}
	\begin{tabular}{|c|c|}\hline
		Symbols & Description\\\hline
	    $\mathcal{D}_k$ & The $k$-th client local dataset\\
	    $d$& The feature dimensions of the data\\
		$\mathcal{K}_t$ & The set of clients selected at round $t$\\
		$\omega$ & The global model\\
		$q$ & The client participation rate ($0<q<1$)\\
		$K$ & The total number of clients\\
		$k$ & The number of clients participating in training\\
		${\mathbf{H}_t}$ & The Hessian information\\
		$\Gamma$ & The diagonal matrix of Hessian information\\
		$p$& The direction of gradient descent\\
		$\eta$ & The learning rate\\
		$\nabla f(\omega )$ & The gradient information\\
		$S_t$& The stochastic batch picked at iteration $t$\\
		$m$ & The memory size\\
		$B_t$ & The Hessian information at round $t$\\
		$\mathcal{B}_t$ & The Hessian approximation information\\
		$J(\omega)$ & $J(\omega)$ is the Jacobian matrix of $f(\omega)$ with respect to $\omega$\\
		$\tau$ & The number of blocks\\
		$B$ & The local min-batch size\\
		$E$ & The number of local epochs\\
		$T$ & The round of iteration\\
		$\omega_{t}^{k}$ & The local model parameter of the $k$-th client at round $t$\\\hline
	\end{tabular}
	\label{tab-n}
\end{table}

\section{Our Algorithm Design}\label{sec-4}
{In this section, we first present our FIM-based approximate distributed optimization algorithm, and then introduce the designed FedOVA training program in detail. Second, the summaries of symbols are presented in Table \ref{tab-n}.}

\subsection{FIM-based Approximate Distributed Optimization Algorithm}
In the distributed learning community, researchers generally use the classic L-BFGS \cite{liu1989limited} optimization algorithm with stochastic batches to update Eq. \eqref{eq-3}. We first review stochastic L-BFGS algorithm. Given an optimization problem with $d$ variables, L-BFGS only needs to store a few vectors of length $d$ to implicitly approximate the Hessian. Therefore, the intrinsic idea within L-BFGS is to utilize the curvature information implied by the vector pairs $(s_t,y_t)$ to help regularize the gradient direction, where ${s_t} = {\omega _{t + 1}} - {\omega _t}$ and ${y_t} = \nabla f({\omega _{t + 1}}) - \nabla f({\omega _t})$. Specifically, the L-BFGS algorithm needs the leverage history state, i.e., the last $m$ (generally $m=10$) elements in $(s_t, y_t)$ to update the current state. In the stochastic L-BFGS algorithm, $y_t$ is replaced by a stochastic version of $B_t^{{S_t}}({\omega _{t + 1}} - {\omega _t})$, i.e.,
\begin{equation}
{y_t} = B_t^{{S_t}}({\omega _{t + 1}} - {\omega _t}),
\end{equation}
\begin{equation}\label{eq-6}
B_t^{{S_t}}\mathop  = \limits^{def} \frac{1}{{|{S_t}|}}\sum\nolimits_{i \in {S_t}} {{\nabla ^2}{f_i}({\omega _t})},
\end{equation}
where $S_t$ is the stochastic batch picked at iteration $t$. Then we can use two-loop recursion based vector-free L-BFGS (VL-BFGS) algorithm which proposed in \cite{chen2014large} to update the vector pairs $(s_t,y_t)$. In this way, we can achieve the necessary goals of designing distributed second-order algorithms mentioned above, i.e., Hessian-free communication and inverse-Hessian-free computation. However, such a method is difficult to stabilize the algorithm, and the current measures to remedy this problem require the batch size to be large enough, which is obviously unrealistic in FEEL \cite{liu2018acceleration}.

{Motivated by the inconvenience mentioned above, in this paper, we propose a low-cost Hessian approximation method, i.e., Fisher information-based Hessian approximation method, to approximate the Hessian matrix efficiently. Specifically, according to the definition of Generalized Gauss-Newton matrix (GGN) \cite{chen2021fast,martens2020new}, Eq. \eqref{eq-6} can be written as follows:}

\begin{equation}\label{eq-7}
{B_t^k \approx {[J(\omega _t^k)]^ \top }{\nabla ^2}f(\omega _t^k)J(\omega _t^k)\mathop  = \limits^{def} \mathcal{B}_t^k,}
\end{equation}
{where $J(\omega _t^k)$ is the Jacobian matrix of $f(\omega)$ with respect to $\omega$ at $\omega _t^k$ and we use $\mathcal{B}_t^k$ to denote the Hessian or Hessian approximation used to smoothen $y_t^k$, i.e., $y_t^k = \mathcal{B}_ks_k$. Since we use a stochastic version of the gradient to update the model, we consider a stochastic batch of $S_k$, thus, Eq. \eqref{eq-7} can be rewritten as: }
\begin{equation}\label{eq-8}
    {\mathcal{B}_t^{{S_k}} = \frac{1}{{|{S_k}|}}\sum\nolimits_{i \in {S_k}} {{{[J(\omega _t^{{S_k}})]}^ \top }{\nabla ^2}{f_i}(\omega _t^{{S_k}})J(\omega _t^{{S_k}})} }.
\end{equation}
Recall that, since the cross-entropy loss we use is negative log-likelihood, it is not difficult to obtain ${\mathbb{E}_{x \in {\mathcal{D}_k}}}[\nabla {f_k}({\omega ^*},x)\nabla {f_k}{({\omega ^*},x)^ \top}]$, where $\omega ^*$ is the true parameter (i.e., the Softmax distribution of the local model) obtained in the form of Fisher information. According to the definition of two equivalent methods for calculating the Fisher information matrix \cite{friedman2001elements,ly2017tutorial}, the above mentioned equation can also be written as ${\mathbb{E}_{x \in {\mathcal{D}_k}}}[{\nabla ^2}{f_k}({\omega ^*},x)]$. {This means we can use $\nabla {f_k}({\omega },x)\nabla {f_k}{({\omega },x)^ \top }$ as an asymptotically unbiased estimation of ${\nabla ^2}{f_k}{(\omega ,x)^ \top }$, since w gradually converges to $\omega^*$ during the training. Thus, our proposed Fisher information-based Hessian approximation method is defined as follows:}

\begin{equation}\label{eq-9}
{\mathcal{B}_t^{{S_k}} = \frac{1}{{|{S_k}|}}\sum\nolimits_{i \in {S_k}} {{{[J(\omega _t^{{S_k}})]}^ \top }\nabla {f_i}(\omega _t^{{S_k}})\nabla {f_i}{{(\omega _t^{{S_k}})}^ \top }J(\omega _t^{{S_k}})} .}
\end{equation}


{Additionally, to significantly reduce the computational complexity and storage while maintaining the accuracy, in DNN we leverage the diagonalization technique \cite{liu2018acceleration} to approximate the Hessian matrix, i.e., $\Gamma=\mathrm{diag}(\nabla {f_i}(\omega _t^{{S_k}})\nabla {f_i}{(\omega _t^{{S_k}})^ \top }) \approx \nabla {f_i}(\omega _t^{{S_k}})\nabla {f_i}{(\omega _t^{{S_k}})^ \top }$. Thus, we only need to store the diagonal elements of FIM and make all other elements zero. Therefore, in the proposed FIM-based L-BFGS algorithm, each client computes the diagonal matrix of FIM locally and uploads it to the server. The server collects and aggregates these FIM-vector products to update the vector pair $(s_t, y_t)$. In particular, the communication cost in each round can be $\mathcal{O}(d)$ (for more analysis details, see below). In this way, the proposed algorithm avoids Hessian communication and inverse Hessian computation.}

{We present our FIM-based Approximate L-BFGS Algorithm, as shown in Algorithm \ref{al-2}. As shown in lines 2 to 4, the server sends the initialized global model to the selected client. Then the server aggregates the FIM uploaded from the local client and utilizes the L-BFGS algorithm to update the pair $(s_t, y_t)$, as shown in lines 5 to 15. In addition, all the local computation steps of the client are presented in the $\mathrm{ClientUpdate}$ function.}

\begin{algorithm}[t!]
\caption{FIM-based Approximate L-BFGS Algorithm} \label{al-2}
\textbf{Input:} $\mathcal{K}$ is the client set with indexed by $k$, $S_t^k$ is the stochastic batch of client $k$ at iteration $t$, integer history size $m>0$.\\
\textbf{Output:} Optimized global model $\omega^*$.\\
\textbf{Server}: 
\begin{algorithmic}[1]
\STATE Choose $H_0$
\FOR{each round $t = 1, \ldots, T$} 
\STATE $\mathcal{K}_{t} \leftarrow$ (Sample a subset of clients from $\mathcal{K}$)
\STATE Server sends the $\omega_t$ to $\mathcal{K}_{t}$  \hfill $// \mathrm{\textbf{Communication}}$
\WHILE{global model no converge}
\STATE {Compute a direction ${p_t} =  - {H_t}\nabla f({\omega _t})$ by using two-loop recursion algorithm \cite{chen2014large}}
\STATE Compute ${\omega _{t + 1}} = {\omega _t} + \eta {p_t}$
\STATE {Update the curvature pairs: ${s_t} = {\omega _{t + 1}} - {\omega _t}$, ${y_t} = (\frac{1}{K}\sum\limits_{k = 1}^K {\mathcal{B}_t^{S_t^k}} )({\omega _{t + 1}} - {\omega _t})$, computed by $\mathrm{ClientUpdate}(k, \omega_t)$ function} \hfill $//\mathrm{\textbf{Computation}}$
\IF{$k>=m$}
\STATE Discard vector pair $(s_{t-m},y_{t-m})$ from memory storage
\ELSE
\STATE Store the vector pair $(s_{t},y_{t})$
\ENDIF
\ENDWHILE
\ENDFOR
\end{algorithmic}
$\mathrm{ClientUpdate}$ $(k, \omega)$:  \hfill$//$ \textbf{Run on client $k$}\\
\begin{algorithmic}[1]
\FOR{all edge clients $k = 1,\ldots,K$ \textbf{in parallel}}
	\FOR{each epoch $e=1, \ldots, E$}
	\FOR {batch $\mathcal{S}_{t}^k \in \mathcal{B}$ }
	\STATE {$\mathcal{B}_t^{S_t^k} = \frac{1}{{|{S_k}|}}\sum\nolimits_{i \in S_t^k} {{{[J(\omega _t^{S_t^k})]}^ \top }\mathrm{diag}(\Gamma )} J(\omega _t^{{S_k}})$}
	\ENDFOR
	\ENDFOR
	\ENDFOR
	\RETURN $\mathcal{B}_t^{S_t^k}$ to server \hfill $//$ $\mathrm{\textbf{Communication}}$
\end{algorithmic}
\end{algorithm}

\begin{algorithm}[!t]
\caption{FedOVA training scheme}
\label{alg:fed-ova}
\textbf{Input}: Client set $\mathcal{K}$, $n$ component models $\{ {\omega ^i}\} _{i = 1}^n$\\
\textbf{Output}: $n$ optimal binary classifiers $\{ {f_1}, \ldots ,{f_n}\} $\\
\textbf{Server}: \\
\begin{algorithmic}[1] 
\STATE {Initialize parameters of all component models $\omega_{0}$}
	\FOR{each round $t = 1, \ldots, T$} 
	\STATE $\mathcal{K}_{t} \leftarrow$ (Sample a subset of clients from $\mathcal{K}$)
	\STATE Send the parameters $w_t^{i}$ to $\mathcal{K}_{t}$
	\FOR{each client $k\in \mathcal{K}_{t}$}
	\STATE $\mathrm{ClientUpdate} (k, \omega^{i}_{t})$ (refer to Algorithm \ref{al-2})
	\ENDFOR	
	\FOR{each component model $\omega^i_t$ in group $P_i$ }
	\STATE Update the model $\omega^i_{t+1}$ by using Algorithm \ref{al-2}
	\ENDFOR
	\ENDFOR
\end{algorithmic}
\end{algorithm}

\begin{figure}[!t]
 \centering
 \includegraphics[width=0.75\linewidth]{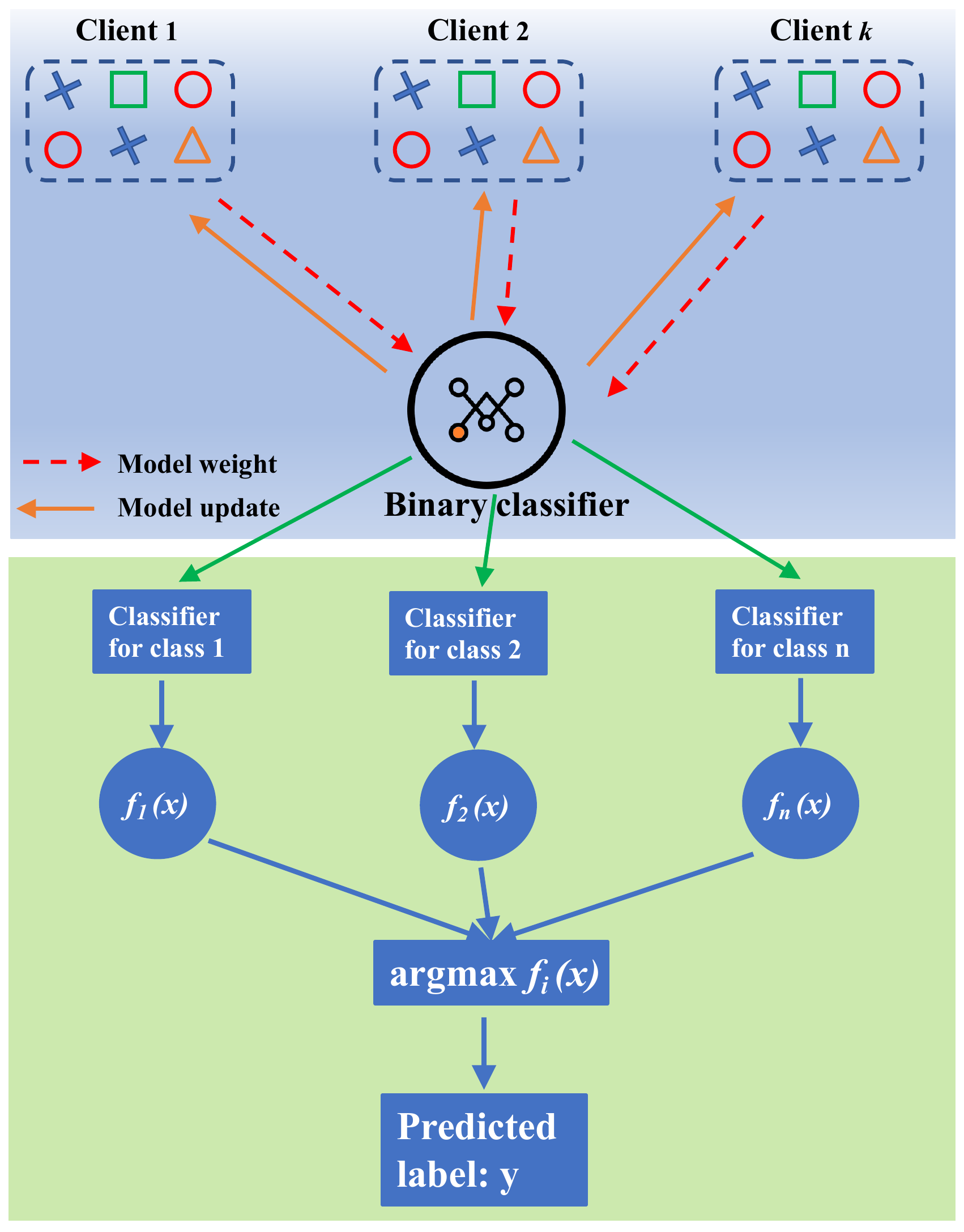}
 \caption{{Overview of the FedOVA training scheme, which trains $n$ binary classifiers and selects the output of the most confident classifier as the prediction result.}}
  \label{fig:overview}
\end{figure}

\subsection{FedOVA Training Scheme}
In this section, we present a detailed training procedure on how to combine OVA and FEEL, i.e., FedOVA (see Algorithm \ref{alg:fed-ova}). We then explain why the proposed FedOVA can address the statistical challenges of federated optimization with non-IID data in a multi-classification task. Recall that the non-IID problem in FEEL means that clients cannot obtain enough labels of multiple classes through data sharing and thus cannot train a high-precision multiclass classification model, thereby limiting the scalability of the FEEL framework. 
We assume that there is a server responsible for the coordination of a client set $\mathcal{K}$ in FEEL, where each client has a local dataset $\mathcal{D}_{k}, k \in \left\{1,~2,~\dots,~K \right\}$ and the total number of classes is $n$. The training procedure of FedOVA repeats the following steps for communication rounds $1$ to $T$:

\noindent \textit{\textbf{Step 1, initialization:}} For each round of training, the server randomly selects a fraction $q$ $(0<q<1)$ of the client set $\mathcal{K}$ to participate in federated edge training, resulting in subset $\mathcal{K}_{t}$. Then, the server broadcasts the binary classifier model parameters $\omega^{i}_{t}$ to the clients in $\mathcal{K}_{t}$, where $i \in  \left\{1,~2,~\dots,~n \right\}$ denotes the classifier ID.

\noindent\textit{\textbf{Step 2, local training:}} 
    {After receiving the binary classifier model parameters $\omega^{i}_{t}$, each client initializes some of the OVA component classifier models according to its own local data label distribution, i.e., $\omega^{i,k}_{t} \leftarrow {\omega^i_{t}}$.} Taking the F-MNIST dataset as an example, if client $k$ only has label ``1'' and label ``2'', then this client initializes the parameters of classifiers $f_1$ and $f_2$. For each binary classifier, the goal is to minimize the following objective function:
    
	\begin{equation}\label{eql:fl_eq1}
    \arg \min \limits_\omega  {\cal L}(\omega ) = \frac{1}{{|{\mathcal{D}_k}|}}\sum\limits_{\{ {x_i},{y_i}\}  \in {\mathcal{D}_k}} \ell  ({y_i},{f_i}({x_i};\omega )),
	\end{equation}
	Each client performs stochastic gradient descent to optimize its classifiers by using the local dataset and then sends the parameter updates back to the server. 
    
\noindent \textit{\textbf{Step 3, aggregation:}} Since each client trains only some of the classifiers, thereby not returning parameters for all of them, and given that all classifiers are independent of one another, we can perform asynchronous updates to reduce the computational burden. {The server groups the returned parameters according to their corresponding binary classifier models and then aggregates the parameters of each group $P_i$. For model $f_i$, the aggregation process can be formulated as:}
\begin{equation}
    \omega_{t+1}^{i}=\frac{1}{|{P}_{t}^{i}|}\sum\limits_{k \in {P}_{t}^{i} } {\omega_{t}^{i,k}}.
\end{equation}
Note that the above steps are repeated until the final ensemble classifier achieves convergence.

\noindent \textbf{Remark:} FedOVA aims to train an ``expert" binary classifier for each class, hence alleviating the problem of multi-classifier model convergence with non-IID data. Combined with FL, conventional OVA can be used to solve non-IID data problems due to the following advantages of FedOVA:

\begin{itemize}

    \item {\textbf{Training is independent.} FedOVA training is independent in the following two aspects: (1) Unlike multi-class classifier ensembles, each classifier in FedOVA is independent and specialized in distinguishing a specific class. Such a design can ensure low error correlation among different classifiers, thus enhancing their collective diversity and improving the overall classification accuracy \cite{Hashemi2009ova}. (2) The execution of FedOVA does not affect the execution of Algorithm \ref{al-2}, so it can be organically integrated with Algorithm \ref{al-2}.}
    
    \item \textbf{Comparable accuracy.} As an ensemble learning scheme, FedOVA is capable of achieving high classification accuracy. \cite{rifkin2004defense} has shown that when its binary classifiers are well tuned, OVA can achieve accuracy on par with any other multi-classification approach. 

    \item \textbf{No requirement for all sample classes.} Since each binary classifier in FedOVA classifies samples as either belonging to the current class or not, missing negative sample classes does not significantly affect classification performance. When new classes emerge, FedOVA just needs to create a new classifier for each, thereby allowing for smooth adaptation to environmental changes during federated training.
    

\end{itemize}

\section{Convergence and Complexity Analysis}
In this section, we conduct a comprehensive analysis of the convergence and complexity of the proposed algorithm. First, we provide convergence theoretical analysis for both strongly convex and non-convex cases. Second, we focus on analyzing the local computing and system communication complexity of the proposed algorithm.

\subsection{Convergence Analysis}
In our settings, we use the stochastic batches of the FIM-based L-BFGS algorithm with a fixed learning rate, so the algorithm can only converge to the neighborhood of optimal point. Thus, we assume that the function $f:{\mathbb{R}^d} \to \mathbb{R}$ is $\Lambda $-Lipschitz continuous or $\Lambda $-smooth, i.e., $\forall i$, we have:
\begin{equation}\label{eq-12}
|\nabla {f}(\omega ') - \nabla {f}(\omega )|| \le \Lambda ||\omega ' - \omega ||,\forall \omega ,\omega ' \in \mathbb{R}^d,
\end{equation}
or equivalently,
\begin{equation}\label{eq-13}
{f}(\omega ') \le {f}(\omega ) + \nabla {f}{(\omega )^T}(\omega ' - \omega ) + \frac{\Lambda }{2}||\omega ' - \omega ||,\forall \omega ,\omega \in \mathbb{R}^d.
\end{equation}

\subsubsection{Strongly Convex Case} 
To analyze the convergence of the strong convex case, we need to make the following assumptions hold:
\begin{assumption}\label{assu-1}
\textit{The function $f:{\mathbb{R}^d} \to \mathbb{R}$ is twice continuously differentiable, $L$-smooth, and $\lambda $-strongly convex,  $\lambda >0$, i.e.,}
\begin{equation}\label{eq-14}
\lambda I \le {\mathcal{B}^S} \le LI,
\end{equation}
\textit{where $I \in \mathbb{R}^d$ and $\mathcal{B}$ is the Hessian (approximation) to stabilize $y_t$. In particular, for all batches $S$ of size $b$, the above equation holds.}
\end{assumption}
\begin{assumption}\label{assu-2}
\textit{For all batches $S$ and $\omega \in \mathbb{R}^d$, $\nabla {f^S}(\omega )$ is an unbiased estimator of the true gradient $\nabla F(\omega )$, i.e.,}
\begin{equation}
\mathbb{E}[\nabla {f^S}(\omega )] = \nabla f(\omega ).
\end{equation}
\end{assumption}

According to Assumption \ref{assu-1} and Eq. \eqref{eq-13}, we have:
\begin{equation}
f(\omega ') \le f(\omega ) + \nabla f{(\omega )^\top}(\omega ' - \omega ) + \frac{\lambda }{2}||\omega ' - \omega ||,\forall \omega ,\omega  \in \mathbb{R}{^d}.
\end{equation}
Based on the above assumptions, we can state the following lemma that the Hessian approximation formulated by Algorithm \ref{al-2} are bounded above and away from zero.
\begin{lemma}\label{lemma-1}
\textit{If Assumptions \ref{assu-1}--\ref{assu-2} hold, then there are two constants $\theta_1$ and $\theta_2$ where $0 < {\theta _1} \le {\theta _2}$, so that the $\mathbf{H}_t$ generated by Algorithm \ref{al-2} satisfies the following equation:}
\begin{equation}
   {\theta _1}I \le {\mathbf{H}_t} \le {\theta _2}I,\forall t \in \{ 0,1, \ldots ,T\} .
\end{equation}
\end{lemma}
{Thanks to Lemma \ref{lemma-1}, we can prove the following theorem without making bounded assumptions on stochastic gradients (i.e., $||\nabla f(\omega )|| \le \epsilon $):}
\begin{theorem}\label{theo-1}
\textit{{Let $\omega^*$ be the minimizer of $f(\omega)$, i.e., ${f^*} = f({\omega ^*}) = \arg \mathop {\min }\limits_{\omega  \in {\mathbb{R}^d}} f(\omega )$. We assume that $\omega_t$ is the model parameter iteratively generated by the  FIM-based stochastic L-BFGS algorithm with a constant learning rate $\alpha  \in (0,\frac{{\lambda {\theta _1}}}{\mu })$ where $\mu  = \theta _2^2(\lambda  + \Lambda \beta (b))\Lambda $. If the assumption \ref{assu-2} holds and $f$ is convex, $\forall t \in \{ 0,1, \ldots ,T\} $, we have:}}
\begin{equation}
\begin{aligned}
  \mathbb{E}[f({\omega _t}) - {f^{^*}}] &\leqslant \{ 1 - [1 - 2\alpha (\lambda {\theta _1} - \alpha \mu \} \frac{{\alpha \theta _2^2\Lambda N}}{{4(\lambda {\theta _1} - \alpha \mu )}} \hfill \\
   &+ {[1 - 2\alpha (\lambda {\theta _1} - \alpha \mu )]^t}[f({\omega _0}) - {f^{^*}}]\\&\mathop  \leqslant \limits \frac{{\alpha \theta _2^2\Lambda N}}{{4(\lambda {\theta _1} - \alpha \mu )}} (T \to \infty), \hfill \\ 
\end{aligned}
\end{equation}
\textit{where $\beta (b) = \frac{{n - b}}{{b(n - 1)}}$ and $N = 2\mathbb{E}[||\nabla {f_i}({\omega ^*})|{|^2}]$.}
\end{theorem}

\subsubsection{Non-convex Case}
Similarly, to analyze the convergence of non-convex cases, we need to make the following assumptions:
\begin{assumption}\label{assu-3}
\textit{The function $f:{\mathbb{R}^d} \to \mathbb{R}$ is twice continuously differentiable and $L$-smooth, i.e.,}
\begin{equation}\label{eq-14}
{\mathcal{B}^S} \le LI,
\end{equation}
\textit{where $I \in \mathbb{R}^d$ and $\mathcal{B}$ is the Hessian (approximation) to stabilize $y_t$. In particular, for all batches $S$ of size $b$, the above equation holds.}
\end{assumption}

\begin{assumption}\label{assu-4}
\textit{For all batches $S$ and $\omega \in \mathbb{R}^d$, the function $f(\omega)$ is bounded below by a scalar ${\hat f}$.}
\end{assumption}

\begin{assumption}\label{assu-5}
\textit{For all batches $S$ and $\omega \in \mathbb{R}^d$, different from the strong convex case, here we need to make the bounded gradient assumption that there are constants $\gamma  \ge 0$ and $\eta >0$ that make ${\mathbb{E}_S}[||\nabla {f^S}(\omega )|{|^2}] \le {\gamma ^2} + \eta ||\nabla F(\omega )|{|^2}$ hold.}
\end{assumption}

\begin{assumption}\label{assu-6}
\textit{For all batches $S$ and $\omega \in \mathbb{R}^b$, $\nabla {f^S}(\omega )$ is an unbiased estimator of the true gradient $\nabla f(\omega )$, i.e.,}
\begin{equation}
\mathbb{E}[\nabla {f^S}(\omega )] = \nabla f(\omega ).
\end{equation}
\end{assumption}

Similar to strongly convex case, we also can state the following lemma that the Hessian approximation formulated by Algorithm \ref{al-2} are bounded above and away from zero.

\begin{lemma}\label{lemma-2}
\textit{If assumptions \ref{assu-3}--\ref{assu-6} hold, then there are two constants $\theta_1$ and $\theta_2$ where $0 < {\theta _1} \le {\theta _2}$, so that the $\mathbf{H}_t$ generated by Algorithm \ref{al-2} satisfies the following equation:}
\begin{equation}
   {\theta _1}I \le {\mathbf{H}_t} \le {\theta _2}I,\forall t \in \{ 0,1, \ldots ,T\} .
\end{equation}
\end{lemma}

Thanks to Lemma \ref{lemma-2}, we can prove the following theorem as follows:
\begin{theorem}\label{theo-2}
\textit{Let $\omega^*$ be the minimizer of $F(\omega)$, i.e., ${F^*} = F({\omega ^*}) = \arg \mathop {\min }\limits_{\omega  \in {\mathbb{R}^d}} F(\omega )$. We assume that $\omega_t$ is the model parameter iteratively generated by the  FIM-based stochastic L-BFGS algorithm with a constant learning rate $\alpha  \in (0,\frac{{{\theta _1}}}{{\theta _2^2\eta \Lambda }})$. If the assumption \ref{assu-3}--\ref{assu-6} hold, $\forall t \in \{ 1, ~2,~\ldots,~T\} $, we have:}

\begin{equation}
    \begin{aligned}
\mathbb{E}[\frac{1}{T}\sum\nolimits_{t = 0}^{T - 1} {||\nabla f({\omega _t})|{|^2}} ] &\leqslant \frac{{\alpha \theta _2^2{\gamma ^2}\Lambda }}{{{\theta _1}}} + \frac{{2[f({\omega _0}) - {f^*}]}}{{\alpha {\theta _1}T}} \hfill \\
   &\leqslant \frac{{\alpha \theta _2^2{\gamma ^2}\Lambda }}{{{\theta _1}}}(T \to \infty ). \hfill \\ 
    \end{aligned}
\end{equation}

\end{theorem}

\subsection{Complexity Analysis}
In our setting, the communication costs of the proposed algorithm involves the evaluation of $\mathcal{B}_t^{{S_t}}$, $\nabla {f^{{S_t}}}(\omega )$, and Algorithm \ref{al-2}. We state that the distributed optimization and communication cost theorem in the proposed system is as follows.

\begin{assumption}\label{assu-7}
{Diagonal Hessian of $\mathbf{H}_t$. The Hessian of the loss function ${\cal L}(\omega )$ with respect to the ${\nabla ^2}f(\omega )$ is always diagonal.}
\end{assumption}

{Since we use the stochastic version of gradient descent to update the model, we consider a stochastic batch $S_k$. In this context, we assume that the batch is split into $\tau$ blocks, where the blocks are represented as ${S_{{k_1}}},{S_{{k_2}}}, \ldots ,{S_{{k_\tau }}}$ and assuming that the corresponding Jacobian block matrix is ${J^{{S_{{k_1}}}}},{J^{{S_{{k_2}}}}}, \ldots ,{J^{{S_{{k_\tau }}}}}$. Based on Assumption \ref{assu-7}, the product of the Hessian vector and any vector $v$ can be written as follows:}

\begin{equation}
\begin{aligned}
{\mathcal{B}_k^{{S_k}}v} &{= \sum\nolimits_{i = 1}^\tau  {{{[J{{({\omega _k})}^{{S_{{k_i}}}}}]}^ \top }{\nabla ^2}f{{({\omega _k})}^{{S_{{k_i}}}}}J({\omega _k})} v} \\
&{= \sum\nolimits_{i = 1}^\tau  {\mathcal{B}_k^{{S_{{k_i}}}}v}}
\end{aligned}
\end{equation}

\begin{theorem} \label{theo-3}
\textit{{If Assumption \ref{assu-7} holds, the total communication cost of each iteration of the proposed FIM-based L-BFGS algorithm is $\mathcal{O}(d\log (\tau) + {m^2})$, where $\tau \geqslant {m^2} + m$ is the number of clients and $m$ is the memory size.}}
\end{theorem}

We prove the above theorem as follows:
\begin{proof}
{First, we analyze the communication costs required for evaluating ${g_t} = \nabla {f^{{S_t}}}({\omega _t})$. The communication cost of the server broadcasting $\omega_t$ to $k$ clients is $\mathcal{O}(d)$ and the communication cost of obtaining the sum of local gradients from the clients is $\mathcal{O}(d\log (\tau))$.} {Second, the clients need to store vector pairs $\{ ({s_t},{y_t})\} ,\{ ({s_t},{g_t})\} $ for each iteration and compute every FIM dot-products defined in Eq. \eqref{eq-8}. To evaluate ${y_t} = \mathcal{B}_t^{{S_{t,k}}}{s_t} = \sum\nolimits_{k = 1}^K {\mathcal{B}_t^{{S_{t,k}}}{s_t}}$, the server needs to broadcast the new $s_t$ to all clients with a communication cost of $\mathcal{O}(d)$ and receive the sum of local computation results ${\mathcal{B}_t^{{S_{t,k}}}{s_t}}$ with a communication cost of $\mathcal{O}(d\log(\tau))$. Then the server broadcasts it again with a communication cost of $\mathcal{O}(d)$. Therefore, the system needs to spend a total of communication cost $\mathcal{O}(d + d\log (\tau) + d) = \mathcal{O}(d\log (\tau))$ to execute the whole procedure.} {Third, after calculating the FIM dot-products locally on the client, the client needs to upload it to the server to execute Algorithm \ref{al-2}, where the communication cost is $\mathcal{O}(m^2+m)=\mathcal{O}(m^2)$. Then, the server needs to perform a linear search on the vector pair $(s_t, y_t)$ to update $y_t$, where the communication cost is $\mathcal{O}(m+d)$. Therefore, the whole procedure has a total communication cost of $\mathcal{O}({m^2} + d + m) = \mathcal{O}({m^2} + d)$.} {Hence, the total communication cost of each iteration of the proposed FIM-based L-BFGS algorithm is:} {$\mathcal{O}(d\log (\tau) + d\log (\tau) + {m^2} + d) = \mathcal{O}(d\log (\tau) + {m^2})$.}
\end{proof}

\noindent\textbf{Comparison with FedAvg-type SGD method:} In FEEL, FedAvg-type GD (or SGD) is a common aggregate approach to update the global model. In this setting, the communication cost of the server broadcasting $\omega_t$ to $k$ clients is $\mathcal{O}(d)$ and the communication cost of obtaining the sum of local gradients from the clients is $\mathcal{O}(kd)$. Therefore, the whole procedure has a total communication cost of $\mathcal{O}(kd+d) = \mathcal{O}(kd)$. For this reason, we can see that the communication complexity of the proposed algorithm is smaller than that of FedAvg-type SGD. In particular, compared to the commonly used methods in FEEL, our method has a linear acceleration, which is friendly to resource-constrained FEEL frameworks.

\section{Experiments}\label{sec-5}
To evaluate the performance of our method under non-IID data settings, we conduct extensive experiments on three representative public datasets. All experiments were developed using Python 3.7 and PyTorch 1.7 \cite{paszke2017automatic}, and were executed on a server with an NVIDIA GeForce RTX2080 Ti GPU and an Intel Xeon Silver 4210 CPU. Note that all experiments were performed sequentially to mimic distributed training.

\subsection{Experimental Setup}
We conduct our experiments on three established datasets: F-MNIST \cite{xiao2017fashionmnist}, CIFAR-10 \cite{krizhevsky2009learning}, and the Speech Commands dataset \cite{warden2018speech}. The first two are image datasets with $10$ classes and have been widely used in FL benchmarks. {The Speech Commands dataset contains $35$ classes of $1$-second audio samples stored in WAV format. We select $10$ common keywords for experimental consistency to generate a KeyWord Spotting (KWS) dataset. Each dataset is split into training and test sets, as detailed below. We then assign training samples to $K=100$ clients according to non-IID configurations and train $n=10$ Convolutional Neural Networks (CNNs) as binary classifiers. By default, we select $20\%$ of clients for training at each round, i.e., $C=0.2$. Each client trains on its local dataset $D_{k}$ for $E=5$ epochs with a batch size of $B=15$. The experimental settings for each dataset are as follows:}

\noindent \textbf{F-MNIST:} F-MNIST consists of $60,000$ training samples and $10,000$ test samples. We employ the architecture in \cite{mcmahan2017communication} using two convolution layers with $16$ and $32$ channels, respectively. Each convolution layer is followed by a $2 \times 2$ max-pooling layer and activated using the ReLU function.

\noindent\textbf{CIFAR-10:} For CIFAR-10, we adopt the convolutional VGG11 architecture \cite{Simonyan15} and distribute the $50,000$ training images to $K$ clients for simulation. 
    
\noindent \textbf{KWS:} We select $10$ keywords (i.e., ``one", ``two", ``three", ``four", ``five", ``down", ``left", ``right", ``stop", ``go") from the Speech Commands dataset to generate our KWS dataset. We extract $50 \times 16$ Mel Frequency Cepstral Coefficients (MFCCs) as features by sampling each audio. We thereby obtain $21,452$ samples, of which $20,000$ are used for training and the rest for testing. We utilize a $3$-hidden-layer CNN architecture with $16$, $32$, and $64$ channels, followed by a fully-connected layer with $256$ units. Each convolution layer is followed by a $1 \times 2$ max pooling layer. All convolution and fully-connected layers are ReLU-activated.

\noindent \textbf{Remark:} {For the non-IID configuration, we use the parameter $1 \leq l \leq 10$ to indicate the number of unique labels held by the client. For instance, non-IID-$2$ means that each client has $2$ distinct labels. This is achieved by grouping the training data by label and dividing each group into $(l \times K)/n$ partitions, finally assigning each client $l$ partitions with different labels.}

\begin{table}[!t]
\centering
\caption{{Performance Comparison with Different Distributed Optimization Algorithms}}
\label{tab:my-table}
\begin{tabular}{@{}cccc@{}}
\toprule
Method                            & Dataset  & Training Round & Accuracy (\%) \\ \midrule
\multirow{3}{*}{FedAvg-based SGD} & F-MNIST  &    200            &  98.2       \\
                                  & CIFAR-10 &   200             &    79.2     \\
                                  & KWS      &   250             &    94.4      \\\midrule
\multirow{3}{*}{FedAvg-based Adam} & F-MNIST  &     200           &  98.1        \\
                                  & CIFAR-10 &    200            &   77.9       \\
                                  & KWS      &    250            &   92.8       \\\midrule
\multirow{3}{*}{FedDANE} & F-MNIST  &  150              &    97.8      \\
                                  & CIFAR-10 &   125             &    75.9      \\
                                  & KWS      &   200             &    94.3      \\\midrule
\multirow{3}{*}{\textbf{Our Method}} & F-MNIST  & \textbf{50}   &  \textbf{97.5}        \\
                                  & CIFAR-10 &   \textbf{75}    &  \textbf{78.6}        \\
                                  & KWS      &     \textbf{100} &  \textbf{93.1}       \\ \midrule
\end{tabular}
\end{table}

\subsection{Performance Comparison with Different Distributed Optimization Algorithms}
{In this experiment, we compare the convergence speed of the proposed second-order optimization algorithm with the commonly used FedDANE \cite{li2019feddane} (Federated Newton-Type Method, second-order), FedAvg-based SGD (first-order), and FedAvg-based Adam (first-order) optimization algorithms. First, we conduct experiments on the F-MNIST, CIFAR-10, and KWS datasets under the IID setting to observe the convergence speed of different optimization algorithms. As shown in Table \ref{tab:my-table}, the proposed algorithm converges faster than the baseline methods on the F-MNIST, CIFAR-10, and KWS datasets under the IID setting. For example, For example, for the KWS dataset, our algorithm converges $2 \times$ as fast as the popular Federal Newton method and $2.5 \times$ faster than the conventional first-order optimization algorithm. The reason is that the gradient information and curvature information are used in the proposed algorithm to find a ``better'' descent direction to speed up the convergence speed.}

However, as shown in Table \ref{tab:my-table}, it can be seen from the experimental results that although the proposed algorithm converges faster than the baseline methods, its performance is slightly lower than that of the baseline methods. This implies that convergence speed and accuracy are a trade-off in the distributed second-order optimization algorithm. In practice, SGD and Adam can use first-order information (i.e., gradient information) to find a better convergence point so that the performance is better than the second-order methods. It is worth noting that such a slight loss of precision in exchange for high communication efficiency is friendly to the resource-constrained FEEL framework. Therefore, in future research, we will focus more on improving the performance of the second-order optimization method.

\subsection{Performance Comparison with Different Robust Training Schemes}

\begin{figure*}[!t]
 \centering
 \includegraphics[scale=0.20]{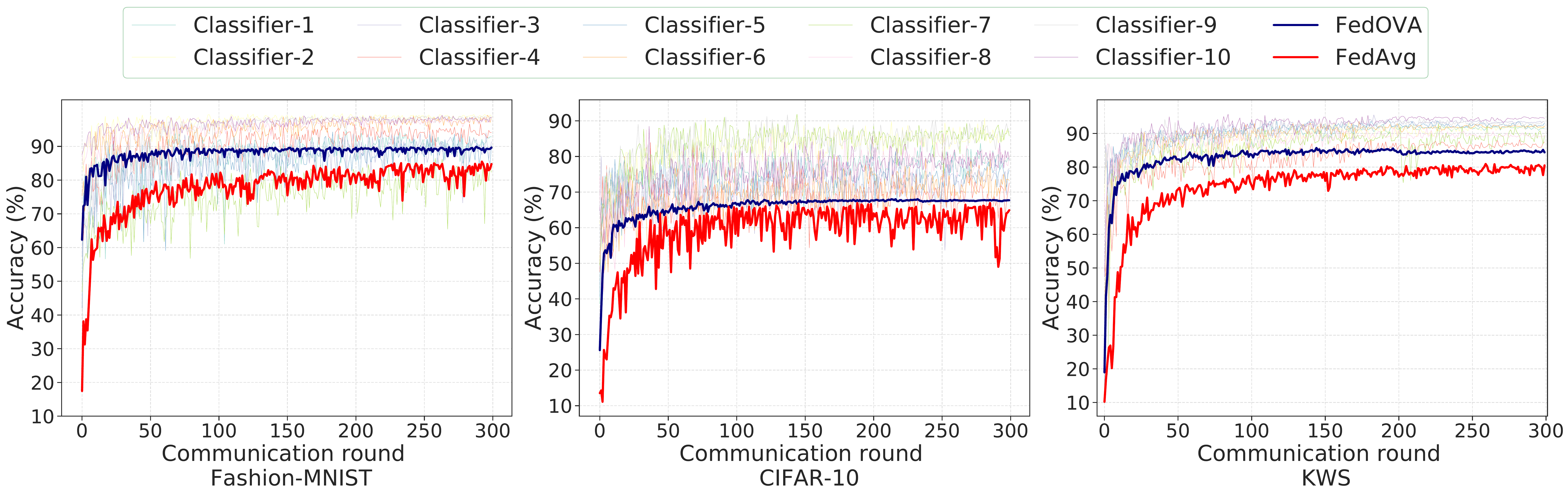}
 \caption{Performance of FedOVA with non-IID data. Each client has samples from only two classes (non-IID-$2$).}
  \label{fig:fedova-overview}
\end{figure*}

\begin{table*}[!t]
\small
\centering
\caption{Accuracy (\%) under different non-IID configurations.}
\label{tab:fedova-noniid-config}
\begin{tabular}{lccccccccc}
\toprule
\multicolumn{1}{l}{Dataset} &\multicolumn{3}{c}{F-MNIST} & \multicolumn{3}{c}{CIFAR-10}& \multicolumn{3}{c}{KWS} \\
\cmidrule(lr){1-1}
\cmidrule(lr){2-4} \cmidrule(lr){5-7} \cmidrule(lr){8-10} \multicolumn{1}{l}{Configuration}
& non-IID-2   & non-IID-3   & non-IID-5 & non-IID-2   & non-IID-3   & non-IID-5 & non-IID-2   & non-IID-3   & non-IID-5      \\
\midrule
FedAvg  & 84.3  & 85.8  & 89.9  & 63.5  & 66.3  & 72.5  & 80.5  & 82.4   & 86.0\\

FedOVA  & \textbf{89.4}  & \textbf{90.3}  & \textbf{91.7}  & \textbf{67.8}  & \textbf{69.1}   & \textbf{73.2} & \textbf{84.6}  & \textbf{86.4}   & \textbf{89.2}  \\
\bottomrule
\end{tabular}
\end{table*}

First, we evaluate the performance of FedOVA with non-IID data against FedAvg.
We adopt the same data distribution method for each dataset and send the training data to $100$ clients according to the non-IID-$2$ configuration for experimental consistency. 
{Our experimental results are shown in Fig.\ref{fig:fedova-overview}. It is evident that FedOVA achieved higher classification accuracy than FedAvg on all three datasets.}
{This is since FedOVA trains $10$ binary classifiers asynchronously, each with high accuracy in distinguishing the current class from other classes.} 
Moreover, the component binary classifiers in FedOVA are independent and have a low error correlation. They are also insensitive to missing samples in some classes. Therefore, an effective multi-class classifier can be constructed by ensembling these component classifiers.

To better investigate the performance of FedOVA under non-IID data distributions, we conduct a series of simulations with different non-IID configurations, i.e., we test non-IID-$l$ for $l\in\{2, 3, 5\}$. 
Our simulation results are presented in Table \ref{tab:fedova-noniid-config}. Using either the FedOVA or the FedAvg algorithm, classification accuracy improved as the number of classes in the client's local dataset increased. 
However, there are still two points worthy of attention: the first is that different non-IID configurations had a dramatic impact on the performance of FedAvg but did not have a significant influence on FedOVA. 
The second is that the FedOVA achieved superior performance compared to FedAvg in every non-IID scenario. These two observations fully illustrate the robustness of our proposed method for non-IID environments.


\begin{table}[!t]
\small
\centering
\caption{Comparison with \protect\cite{zhao2018federated} in terms of classification accuracy (\%) under different data sharing rates $\beta$.}
\label{tab:comp-sharing}
\begin{tabular}{lccc}
\toprule
Dataset & F-MNIST   & CIFAR-10  & KWS \\ 
\midrule
Data sharing $(\beta=5\%)$   & 86.3  & 65.3  & 82.6     \\
Data sharing $(\beta=10\%)$  & 88.1  & 67.5  & 83.5      \\
FedOVA     & \textbf{89.4}  & \textbf{67.8}      & \textbf{84.6}      \\ 
\bottomrule
\end{tabular}
\end{table}

Second, we compare FedOVA with the data-sharing approach in \cite{zhao2018federated}, which is one of the most popular methods for performing federated training with non-IID data.
We set the number of clients $K=100$, $C=0.2$, and use non-IID-$2$ as our experimental configuration. We define $\beta=D_{share}/D_{k}$ as the sharing rate, which represents the ratio of the data shared by the server to the local data. 
We randomly sample the global dataset to get the shared dataset $D_s$ and send it to each client, where $D_s=D_k \times \beta$. 
We conduct simulations to compare the performance of the data-sharing strategy and FedOVA under different sharing rate $ \beta $ settings.
As shown in Table \ref{tab:comp-sharing}, our approach still outperformed the data sharing strategy at a sharing rate of $5\%$ and $10\%$. Only when the server has an extensive global dataset and shares more data with the client can the performance of data sharing outperform FedOVA.
However, increased data sharing leads to a higher risk of privacy leakage, clearly not complying with privacy-preserving requirements. 
Furthermore, if $\beta$ is too large, the generalization performance of the model is undermined. In contrast, our approach does not require sharing any data and provides complete privacy protection.

\subsection{Effect of Large Scale Clients and Small Scale Samples}

Considering the large number of clients that are commonly encountered in FEEL, we evaluate the performance of the FedOVA algorithm with a large-scale setting and compare it with that of FedAvg. 
We hold the total number of training samples constant and significantly increase the number of clients (i.e., $K$ increases from $100$ to $1000$ in F-MNIST and CIFAR-10; since the amount of KWS data is limited in comparison, we set $K=500$), so the amount of local data per client is reduced accordingly.
For each round of training, we set $C=0.2$ and select $K \times C$ clients to participate. We then set the average of the last $20$ rounds as the final accuracy. 
The experimental results are summarized in Table \ref{tab:large}. It appears that the use of FedAvg led to a substantial decrease in accuracy as the number of clients increased; crucially, this was the case with FedOVA.
{Our method adopts the idea of OVA to train several binary classifiers, and the increase in the number of clients also improves the robustness of the binary classifier to different environments. In this way, it is still possible to achieve competitive results with fewer data.}

\begin{table}[!t]
\centering
\small
\caption{Accuracy (\%) vs. the number of clients $K$.}
\label{tab:large}
\begin{tabular}{lcccccc}
\toprule
\multicolumn{1}{l}{Dataset} & \multicolumn{2}{c}{F-MNIST} & \multicolumn{2}{c}{CIFAR-10}& \multicolumn{2}{c}{KWS} \\

\cmidrule(lr){1-1} \cmidrule(lr){2-3}  \cmidrule(lr){4-5} \cmidrule(lr){6-7} 
\multicolumn{1}{l}{K}
&  100 &  1000  &  100 &  1000  &  100 &  500 \\
\midrule
FedAvg  & 84.3  & 83.1  & 63.5  &  61.2  & 80.5   & 75.9 \\ 
FedOVA  & \textbf{89.4}  & \textbf{88.9}  & \textbf{67.8}  & \textbf{66.3}  & \textbf{84.6}   & \textbf{82.4}  \\
\bottomrule
\end{tabular}
\end{table}

\subsection{Effect of Hyperparameter}

\begin{figure}[t]
	\centering
	\subfigure [Accuracy vs. batch size.]{\includegraphics[width=0.45\linewidth]{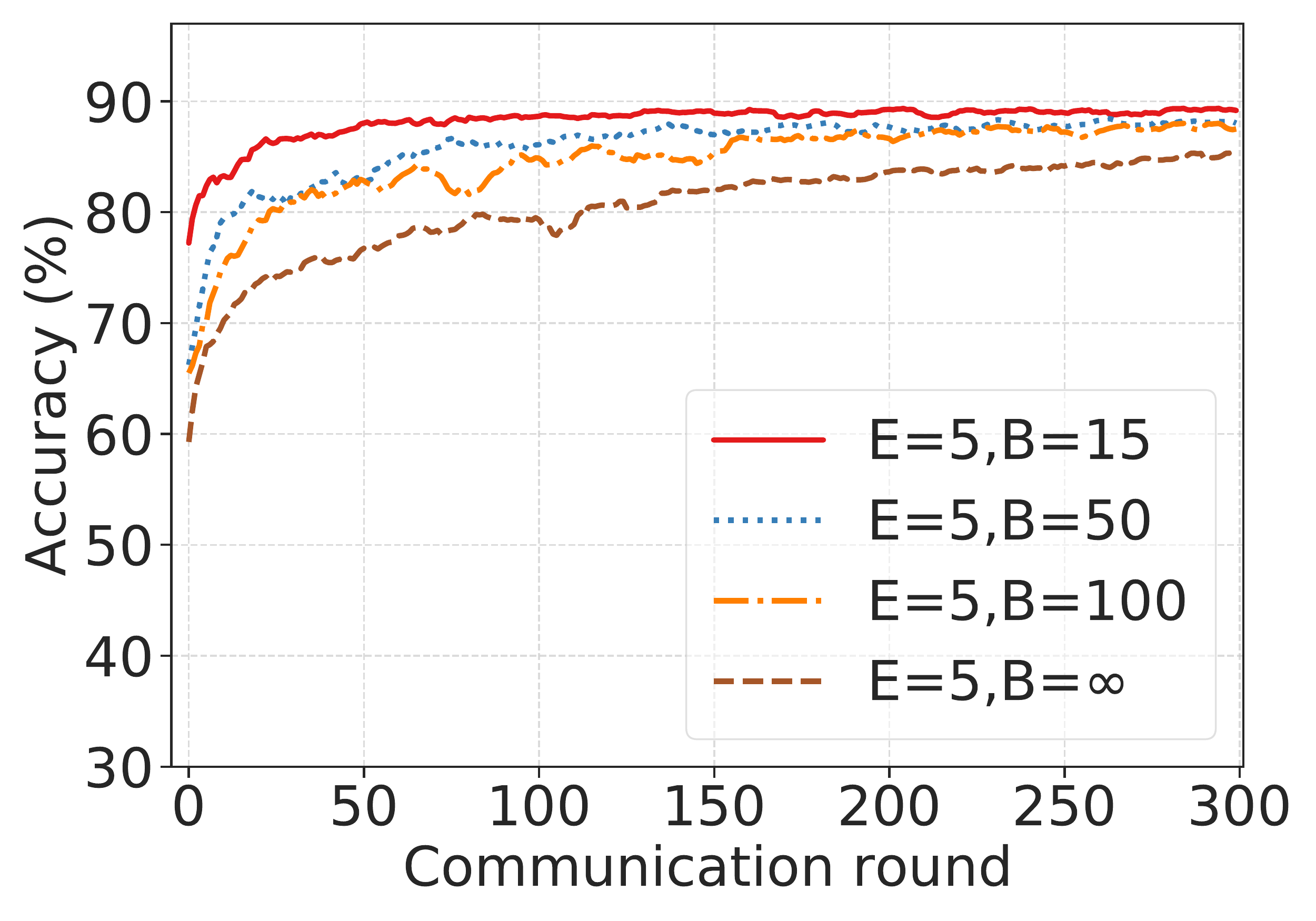}
		\label{fig:fedova-para1}}
	\subfigure[Accuracy vs. the number of training epochs.]{	\includegraphics[width=0.45\linewidth]{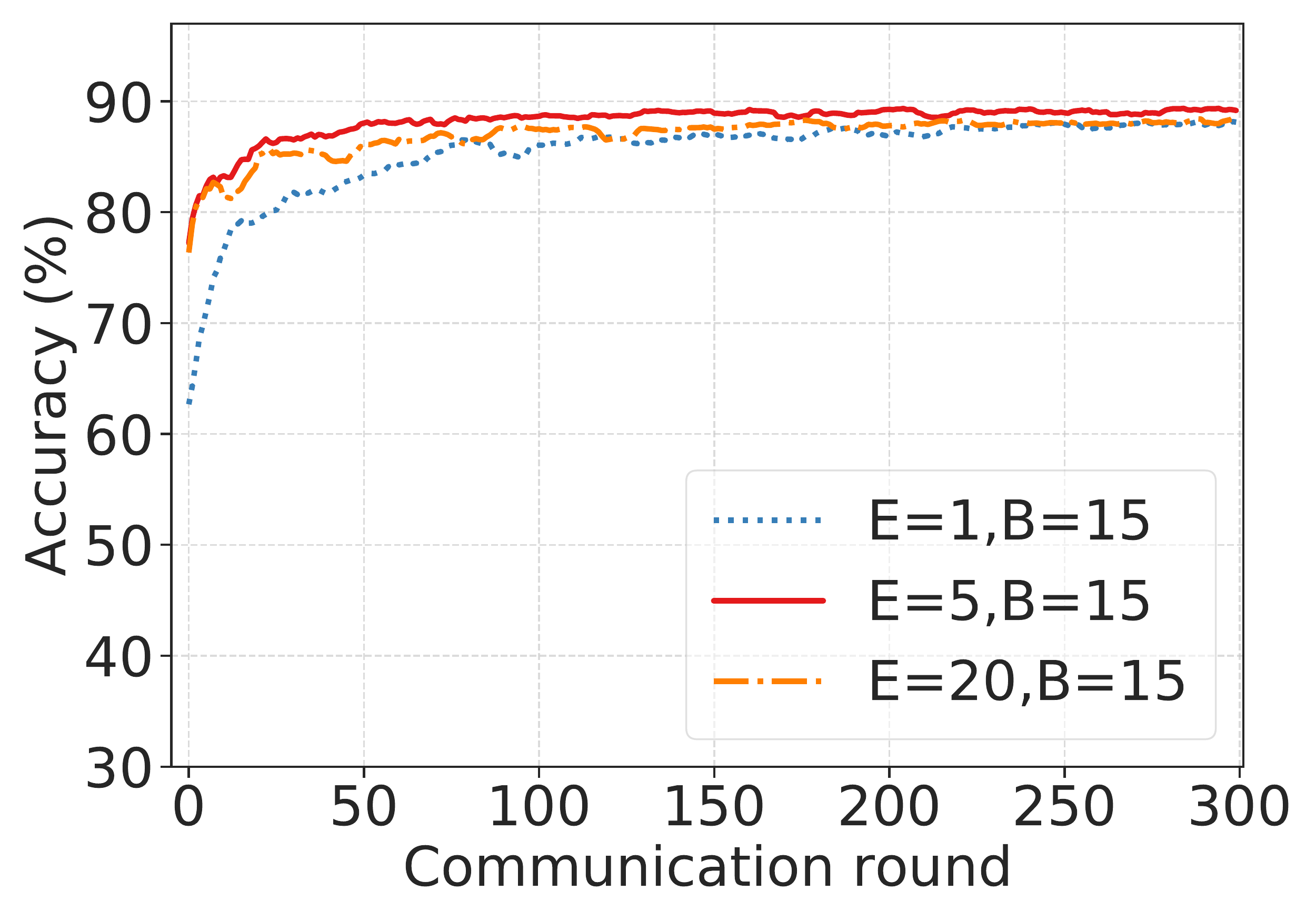}
		\label{fig:fedova-para2}}
	\caption{Performance of FedOVA when varying either the number of epochs $E$ or batch size $B$.}
\label{fig:fedova-component}
\end{figure}

In this subsection, we investigate the effect of different hyperparameter settings on classification accuracy to empirically validate the feasibility of our algorithm in cases that resemble real-world scenarios. 
We conduct experiments on the F-MNIST dataset to study the impact on classification accuracy and training convergence under different training epochs $E$ and batch sizes $B$. 
For illustration purposes, we smoothen the curve and set the accuracy of the next $10$ rounds as the value of the curve on each communication round.

{One may observe in Fig. \ref{fig:fedova-para1} that when fixing the number of training epochs $E=5$, FedOVA achieved similar accuracy rates for different batch sizes ($B=15$, $50$, $100$, $\infty$; note that $\infty$ means training on all the data in a single epoch). However, as the batch size in each epoch increased, the convergence of the algorithm slowed down.}
This result indicates that performing more gradient descent updates during each training round helps accelerate model convergence, enabling the model to reach a satisfactory performance in a short period of time.
{The results in Fig. \ref{fig:fedova-para2} further verify this point. When we reduced the number training epochs, there was also a decrease in the convergence speed of the model in the same period of time. The above results are also in agreement with \cite{mcmahan2017communication}.}


\section{Conclusion}\label{sec-6}
{This paper proposed a promising solution to simultaneously alleviate expensive communication costs and heterogeneous data problems for resource-constrained FEEL. Our insight is to design a robust training scheme for heterogeneous data and does not conflict with efficient communication algorithms. To improve the communication efficiency of resource-constrained FEEL, we have customized an FIM-based distributed approximate second-order optimization algorithm. Furthermore, we proved that the algorithm has linear convergence in the case of strong convexity and non-convexity, and analyzed its computational and communication complexity. Besides, to design an algorithm coupled with the above-mentioned communication efficient algorithm and robust to heterogeneous data, we first proposed a federated training scheme named FedOVA with the help of the traditional OVA method. With extensive experiments, we effectively validate the expected properties of our algorithm and empirically demonstrate its capability of reducing the communication cost and being robust to heterogeneous data.}

In future work, we will further analyze the characteristics of each component classifier in FedOVA in an attempt to find an optimal training approach towards improving the performance of each model. Furthermore, we will investigate how to combine this scheme with asynchronous training to obtain better adaptation to real scenarios.

\bibliographystyle{IEEEtran}
\bibliography{ijcai21}

\begin{IEEEbiography}[{\includegraphics[width=1in,height=1.25in,clip,keepaspectratio]{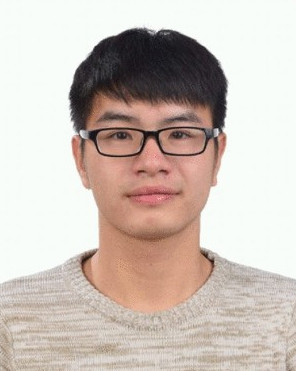}}]{Yi Liu (S'19)}
received the B.Eng degree from the Heilongjiang University, China, in 2019. His research interests include security and privacy in smart city and edge computing, deep learning, intelligent transportation systems, and federated learning.
\end{IEEEbiography}

\begin{IEEEbiography}[{\includegraphics[width=1in,height=1.25in,clip,keepaspectratio]{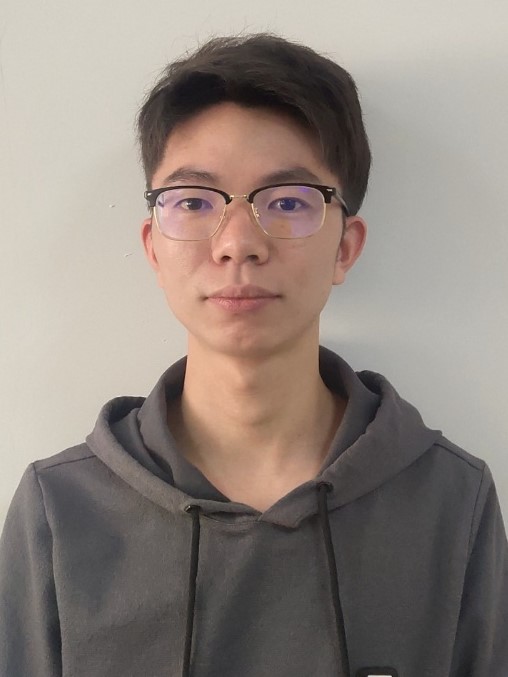}}]{Yuanshao Zhu (S'20)}
received the B.Eng. degree in telecommunication engineering from Shandong University, Weihai, China, in 2019. He is currently a Master Student with the Department of Computer Science and Engineering, Southern University of Science and Technology, Shenzhen, China. His research interests include deep learning in smart city and edge computing, intelligent transportation systems, and federated learning.
\end{IEEEbiography}

\begin{IEEEbiography}[{\includegraphics[width=1in,height=1.25in,clip,keepaspectratio]{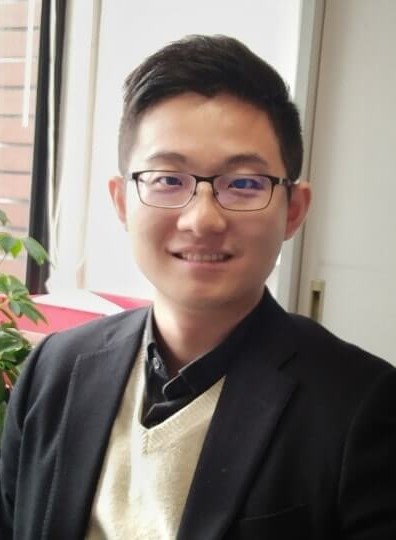}}]{James J. Q. Yu} (S'11--M'15--SM'20) is an assistant professor at the Department of Computer Science and Engineering, Southern University of Science and Technology, Shenzhen, China, and an honorary assistant professor at the Department of Electrical and Electronic Engineering, the University of Hong Kong. He received the B.Eng. and Ph.D. degree in electrical and electronic engineering from the University of Hong Kong, Pokfulam, Hong Kong, in 2011 and 2015, respectively. He was a post-doctoral fellow at the University of Hong Kong from 2015 to 2018. He currently also serves as the chief research consultant of GWGrid Inc., Zhuhai, and Fano Labs, Hong Kong. His general research interests are in smart city and urban computing, deep learning, intelligent transportation systems, and smart energy systems. His work is now mainly on forecasting and decision making of future transportation systems and basic artificial intelligence techniques for industrial applications. He was ranked World's Top 2\% Scientists by Stanford University in 2020. He is an Editor of the \textsc{IET Smart Cities} journal and a Senior Member of IEEE. \end{IEEEbiography}

\clearpage
\appendices
\section{Proofs}
We provide proofs for the theorems and lemmas.

\subsection{Proof of Lemma \ref{lemma-1}}\label{proof-1}
\begin{proof}
Let $\mathcal{H}_t$ be the Hessian approximation, i.e., ${\mathcal{H}_t} = {\mathbf{H}_t}^{ - 1}$. In the L-BFGS algorithm, the iterative update of Hessian is as follows (note that the superscript $(i)$ of $\mathcal{H}_t$ represents the iteration of $m$ Hessian updates in each iteration), i.e., for $i = 0,1, \ldots, m - 1,j = t - m + 1 + i$, we have:
\begin{equation}\label{eq-23}
{\mathcal{H}_t}^{(i + 1)} = \mathcal{H}_t^{(i)} - \frac{{\mathcal{H}_t^{(i)}{s_j}s_j^\top\mathcal{H}_t^{(i)}}}{{s_j^T\mathcal{H}_t^{(i)}{s_j}}} + \frac{{{y_j}y_j^T}}{{y_j^T{s_j}}}.
\end{equation}
In Eq. \eqref{eq-23}, we set ${\mathcal{H}_t} = \mathcal{B}_t^{(m)}$. In FIM-based L-BFGS algorithm, $\mathcal{B}_t = {\nabla ^2}{f^{{S_t}}}({\omega _t})$ is symmetric, i.e., $\mathcal{B}_t^\top = {\mathcal{B}_t}$, thus, we have:
\begin{equation}
||{y_t}|{|^2} = y_t^T{y_t} = s_t^\top\mathcal{B}_t^\top{\mathcal{B}_t}{s_t},
\end{equation}
and by Assumption \ref{assu-1}, we have:
\begin{equation}
\begin{gathered}
  s_t^T\mathcal{B}_t^T{\mathcal{B}_t}{s_t} \geqslant {\theta _1}s_t^T\mathcal{B}_t^T{s_t} = {\theta _1}y_t^T{s_t}, \hfill \\
   \Rightarrow s_t^T\mathcal{B}_t^T{\mathcal{B}_t}{s_t} \leqslant {\theta _2}s_t^T\mathcal{B}_t^T{\mathcal{B}_t}{s_t} = {\theta _2}y_t^T{s_t}. \hfill \\ 
\end{gathered} 
\end{equation}
Therefore,
\begin{equation}
\Rightarrow {\theta _1} \leqslant \frac{{||{y_t}|{|^2}}}{{y_t^T{s_t}}} \leqslant {\theta _2}.
\end{equation}
\end{proof}

\subsection{Proof of Lemma \ref{lemma-2}}
\begin{proof}
According to the proof of Lemma \ref{lemma-1}, we have:
\begin{equation}
||{y_t}|{|^2} = s_t^T\mathcal{B}_t^T{\mathcal{B}_t}{s_t} \leqslant {\theta _2}s_t^TB_t^T{s_t} = {\theta _2}y_t^T{s_t},
\end{equation}
\begin{equation}
\Rightarrow \frac{{||{y_t}|{|^2}}}{{y_t^T{s_t}}} \leqslant {\theta _2}.
\end{equation}

Furthermore, we have the following Equations hold:
\begin{equation}
{\theta _1}||{s_t}|{|^2} \leqslant y_t^T{s_t} \leqslant ||{y_t}||||{s_t}||,
\end{equation}
\begin{equation}
 \Rightarrow ||{s_t}|| \leqslant \frac{1}{{{\theta _1}}}||{y_t}||,
\end{equation}
thus, we have:
\begin{equation}
y_t^T{s_t} \leqslant ||{y_t}||||{s_t}|| \leqslant \frac{1}{{{\theta _1}}}||{y_t}|{|^2},
\end{equation}
\begin{equation}
    \Rightarrow \frac{{||{y_t}||}}{{y_t^T{s_t}}} \geqslant {\theta _1}.
\end{equation}
Therefore, we can obtain the following Equation:
\begin{equation}
\Rightarrow {\theta _1} \leqslant \frac{{||{y_t}|{|^2}}}{{y_t^T{s_t}}} \leqslant {\theta _2}.
\end{equation}
\end{proof}

\subsection{Proof of Theorem \ref{theo-1}}
\begin{proof}
{According to the proof of Lemma \ref{lemma-1}, we have:}
\begin{equation}
\begin{aligned}
  {f({\omega _{k + 1}})} &{= f({\omega _k} - {\alpha _k}{\mathcal{H}_k}\nabla {f^{{S_k}}}({\omega _k}))} \hfill \\
   &{\leqslant f({\omega _k}) - {\alpha _k}\nabla f({\omega _k})^\top{\mathcal{H}_k}\nabla {f^{{S_k}}}{({\omega _k}) } }\\&{+ \frac{{{\alpha _k}^2\theta _2^2\Lambda }}{2}||\nabla {f^{{S_k}}}({\omega _k})|{|^2}.} \hfill \\ 
\end{aligned}
\end{equation}
{Then, we have:}
\begin{equation}
    \begin{aligned}
 {\mathbb{E}[f({\omega _{k+1}}) - {f^*}]} &{\leqslant f({\omega _k}) - {f^*} - {\alpha _k}\nabla f{({\omega _k})^ \top }{\mathcal{H}_k}\mathbb{E}[\nabla {f^{{S_k}}}({\omega _k})] }\hfill \\
   &{+ \frac{{{\alpha _k}^2\theta _2^2\Lambda }}{2}{\mathbb{E}^{{S_k}}}[||\nabla {f^{{S_k}}}({\omega _k})|{|^2}]} \hfill \\
   &{\leqslant [1 - 2{\alpha _k}(\lambda {\theta _1} - {\alpha _k}\theta _2^2(\lambda  + \Lambda \beta (b)\kappa  )\Lambda )](f({\omega _k}) - {f^*}) }\\&{+ \frac{{{\alpha _k}^2\theta _2^2\Lambda N}}{2}.} \hfill \\ 
    \end{aligned}
\end{equation}
{Since we consider the strong convex case, we have $\nabla f({\omega ^*}) = 0$, thus, $f({\omega _k}) \leqslant f(\omega ) + \nabla f{({\omega _k})^ \top }({\omega ^*} - \omega ) + \frac{\Lambda }{2}||\nabla f({\omega ^*}) - \nabla f(\omega )|{|^2}.$ Therefore, if we use a constant learning rate ${\alpha _k} = \alpha  > 0$, thus, let $o = 1 - 2{\alpha _k}(\lambda {\theta _1} - {\alpha _k}\theta _2^2(\lambda  + \Lambda \beta (b)\kappa )\Lambda )$, we have:}
\begin{equation}
    \begin{aligned}
 { \mathbb{E}[f({\omega _{k + 1}}) - {f^*}] - \frac{{\alpha \theta _2^2\Lambda N}}{{4[\lambda {\theta _1} - \alpha \theta _2^2(\lambda  + \Lambda \beta (b)\kappa )\Lambda ]}}} \hfill \\
 { \leqslant o[(f({\omega _k}) - {f^*}) - \frac{{\alpha \theta _2^2\Lambda N}}{{4[\lambda {\theta _1} - \alpha \theta _2^2(\lambda  + \Lambda \beta (b)\kappa )\Lambda ]}}],} \hfill \\ 
    \end{aligned}
\end{equation}
{According to the above Equations, we can iteratively obtain:}
\begin{equation}
    \begin{aligned}
 { \mathbb{E}[f({\omega _k}) - {f^*}] \leqslant {o^k}[f({\omega _0}) - {f^*}] }\hfill \\
{    + (1 - {o^k})\frac{{\alpha \theta _2^2\Lambda N}}{{4[\lambda {\theta _1} - \alpha \theta _2^2(\lambda  + \Lambda \beta (b)\kappa)\Lambda ]}},} \hfill \\ 
    \end{aligned}
\end{equation}
{so we need the learning rate to satisfy:}
\begin{equation}
   { 0 < o < 1,}
\end{equation}
\begin{equation}
  {\Rightarrow \alpha  \in (0,\frac{{\lambda {\theta _1}}}{{\theta _2^2(\lambda  + \Lambda \beta (b))\Lambda }}),\mu  = \theta _2^2(\lambda  + \Lambda \beta (b))\Lambda }
\end{equation}
\begin{equation}
    { \Rightarrow \alpha  \in (0,\frac{{\lambda {\theta _1}}}{\mu }).}
\end{equation}
\end{proof}

\subsection{Proof of Theorem \ref{theo-2}}
More details of the proof can be found in \cite{berahas2016multi}.

\end{document}